\documentclass[letterpaper,10pt,conference,twoside]{IEEEtran}
\IEEEoverridecommandlockouts

\usepackage[inline]{enumitem}
\usepackage{amsmath,amsfonts}
\usepackage{amssymb}
\usepackage{algorithmic}
\usepackage{algorithm}
\usepackage{array}
\usepackage{textcomp}
\usepackage{stfloats}
\usepackage{url}
\usepackage{verbatim}
\usepackage{graphicx}
\usepackage{tikz}
\usepackage[font=footnotesize]{caption}
\usepackage[font=footnotesize]{subcaption}
\usepackage[noadjust]{cite}

\usepackage{url}

\makeatletter
\let\NAT@parse\undefined
\makeatother
\usepackage[pagebackref=true,breaklinks=true,colorlinks,bookmarks=false]{hyperref}
\makeatletter
\newcommand*{\textlabel}[2]{%
  \edef\@currentlabel{#1}
  \phantomsection
  #1\label{#2}
}
\makeatother

\usepackage{textpos}
\usepackage{amsthm}
\usepackage{xcolor}
\colorlet{RED}{red}

\usepackage{tikz}
\usepackage[scaled]{helvet}
\usepackage{flushend}

\theoremstyle{definition}

\newtheorem{defn}{Definition}[section]

\newtheorem*{pb}{Problem}

\DeclareMathOperator*{\argmax}{arg\,max}

\makeatletter
\newcommand\notsotiny{\@setfontsize\notsotiny\@vipt\@viipt}
\makeatother

\makeatletter
\newcommand\footnoteref[1]{\protected@xdef\@thefnmark{\ref{#1}}\@footnotemark}
\makeatother

\begin{document}
\bstctlcite{IEEEexample:BSTcontrol}

\title{\LARGE\bf RB5 Low-Cost Explorer: Implementing Autonomous Long-Term Exploration on Low-Cost Robotic Hardware}

\author{Adam Seewald${}^\text{1}$, Marvin Chanc{\'a}n${}^\text{1}$, Connor M. McCann${}^\text{2}$, Seonghoon Noh${}^\text{1}$, Omeed Fallahi${}^\text{1}$, Hector Castillo${}^\text{1}$,\\ 
Ian Abraham${}^\text{1}$, and Aaron M. Dollar${}^\text{1}$
  \thanks{This work was partly supported by Yale University and a gift from the Boston Dynamics AI Institute.}
  \thanks{${}^\text{1}$A.\hspace*{.4ex}S., M.\hspace*{.4ex}C., S.\hspace*{.4ex}N., O.\hspace*{.4ex}F., H.\hspace*{.4ex}C, I.\hspace*{.4ex}A., and A.\hspace*{.4ex}M.\hspace*{.4ex}D. are with the Department of Mechanical Engineering and Materials Science, Yale University, CT, USA. Email: {\tt\footnotesize \href{mailto:adam.seewald@yale.edu}{adam.seewald@yale.edu};}}
  \thanks{${}^\text{2}$C.\hspace*{.4ex}M.\hspace*{.4ex}M. is with the School of Engineering and Applied Sciences, Harvard University, MA, USA, but the work was performed while affiliated with Yale University.}
}

\maketitle

\begin{abstract} 
This systems paper presents the implementation and design of RB5, a wheeled robot for 
autonomous 
long-term exploration with fewer and cheaper sensors. 
Requiring just an RGB-D camera and low-power computing hardware, 
the system 
consists of an experimental platform with rocker-bogie suspension. It operates 
in unknown and GPS-denied environments and on indoor and outdoor 
terrains. 
The exploration 
consists of a 
methodology that extends frontier- and sampling-based exploration 
with a path-following vector field and 
a state-of-the-art SLAM 
algorithm. 
The methodology allows the robot to explore its surroundings at lower update frequencies, 
enabling the use of lower-performing and lower-cost hardware while still retaining good autonomous performance.
The approach further consists of a methodology to interact with a remotely located human operator based on an inexpensive long-range and low-power communication technology from the internet-of-things domain (i.e., LoRa) 
and a customized communication protocol. 
The results and the feasibility analysis 
show the possible applications and limitations of the approach. 
\end{abstract}
\vspace*{.9ex}
{\small\bfseries\textit{Code---}The open-source software stack is made available on the project repository webpage\footnoteref{link}.}


\footnotetext[2]{\label{link}{\tt\footnotesize \href{https://github.com/adamseew/rb5}{github.com/adamseew/rb5}}}
\vspace*{.3ex}

\section{Introduction}
\noindent
The promise of autonomous long-term robotic exploration is currently being restricted 
in part by the expense of the required sensing, computing, and mechanical hardware. 
This cost is related to the computational intensity of most common navigation and communication approaches~\cite{lluvia2021active,placed2022survey}, which significantly increases for 
outdoor terrains. Addressing this challenge, we introduce 
techniques to reduce update frequencies and enhance the communication capabilities of existing approaches. By loosening the required update frequencies and communication requirements, our methods enable the use of lower-performing and lower-cost hardware while still retaining good autonomous performance.

Recent efforts in this direction include low-cost robots that, e.g., exploit sensing capabilities of commercial smartphones~\cite{muller2021openbot,zhou2021smartphone} but 
lack crucial components for autonomous long-term exploration such as 
terrain adaptability~\cite{muller2021openbot,srinivasa2019mushr}, 
outdoor navigation~\cite{zhou2021smartphone,faisal2021low}, etc. 
%
Furthermore, in areas that are 
challenging to traverse, 
state-of-the-art approaches rely on humans for supervision and high-level decision-making~\cite{tranzatto2022cerberus,roucek2020darpa,tabib2022autonomous}. 
As a result, robots often operate close to humans or require expensive network equipment such as a mesh of communication devices~\cite{kulkarni2022autonomous,ebadi2020lamp} and existing network infrastructure~\cite{khairuldanial2019mobile,
voigtlander20175g}, thereby restricting autonomous exploration to indoor settings only~\cite{delgado2022oros,
cadena2016past,eldemiry2022autonomous,corah2019communication
}.
%
Our methodology exploits LoRa -- an inexpensive long-range and low-power communication technology~\cite{shanmuga2020survey} from the
internet-of-things domain 
-- and a customized communication protocol. 
This allows a human to intervene when 
the robot 
is unable to move with the local sensory~information. 

\begin{figure}
\begingroup%
  \makeatletter%
  \providecommand\color[2][]{%
    \errmessage{(Inkscape) Color is used for the text in Inkscape, but the package 'color.sty' is not loaded}%
    \renewcommand\color[2][]{}%
  }%
  \providecommand\transparent[1]{%
    \errmessage{(Inkscape) Transparency is used (non-zero) for the text in Inkscape, but the package 'transparent.sty' is not loaded}%
    \renewcommand\transparent[1]{}%
  }%
  \providecommand\rotatebox[2]{#2}%
  \newcommand*\fsize{\dimexpr\f@size pt\relax}%
  \newcommand*\lineheight[1]{\fontsize{\fsize}{#1\fsize}\selectfont}%
  \ifx\svgwidth\undefined%
    \setlength{\unitlength}{247bp}%
    \ifx\svgscale\undefined%
      \relax%
    \else%
      \setlength{\unitlength}{\unitlength * \real{\svgscale}}%
    \fi%
  \else%
    \setlength{\unitlength}{\svgwidth}%
  \fi%
  \global\let\svgwidth\undefined%
  \global\let\svgscale\undefined%
  \makeatother%
  \begin{picture}(1,0.65934064)%
    \lineheight{1}%
    \setlength\tabcolsep{0pt}%
    \put(0,0){\includegraphics[width=\unitlength,page=1]{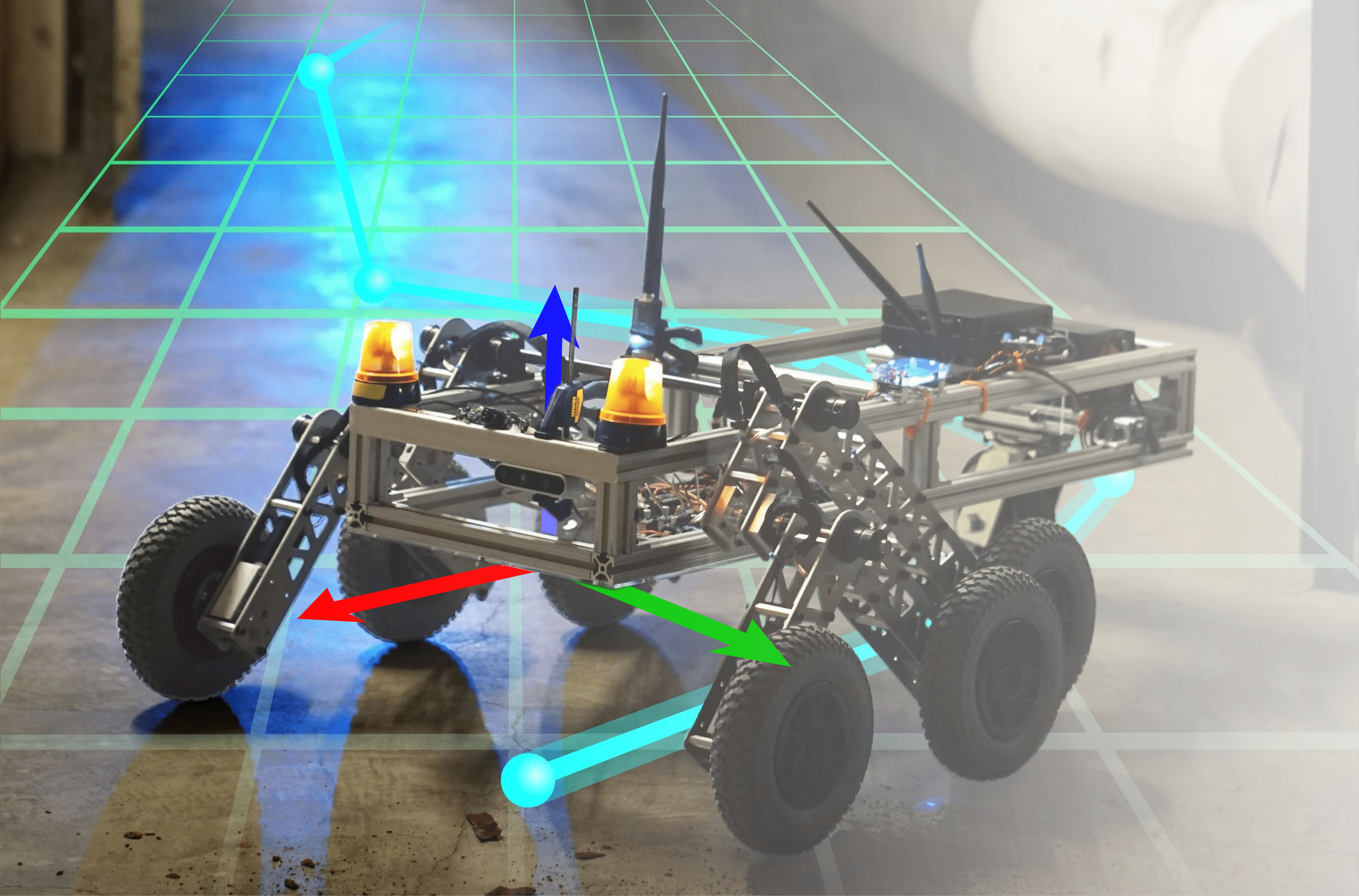}}%
    \put(0.33525123,0.20056237){\makebox(0,0)[lt]{\lineheight{1.25}\smash{\begin{tabular}[t]{l}\color{black}\footnotesize $\mathbf{p}(t_1)$\end{tabular}}}}%
    \put(0.33825123,0.20356237){\makebox(0,0)[lt]{\lineheight{1.25}\smash{\begin{tabular}[t]{l}\color{white}\footnotesize $\mathbf{p}(t_1)$\end{tabular}}}}%
    \put(0.05700171,0.06331295){\makebox(0,0)[lt]{\lineheight{1.25}\smash{\begin{tabular}[t]{l}\color{black}\footnotesize $\mathcal{O}_W$\end{tabular}}}}%
    \put(0.06000171,0.06631295){\makebox(0,0)[lt]{\lineheight{1.25}\smash{\begin{tabular}[t]{l}\color{white}\footnotesize $\mathcal{O}_W$\end{tabular}}}}%

  \end{picture}%
\endgroup%
 
   \caption[RB5 low-cost wheeled robotic explorer]{
     \textbf{RB5 low-cost wheeled robotic explorer}. 
     A robot needs to explore its surroundings with fewer and cheaper sensors -- 
     the picture illustrates RB5, our experimental wheeled robotic platform that carries an RGB-D camera and low-power computing hardware to derive an exploratory coverage path. 
   }
   \vspace*{-.2cm}
   \label{fig:0}
 \end{figure}
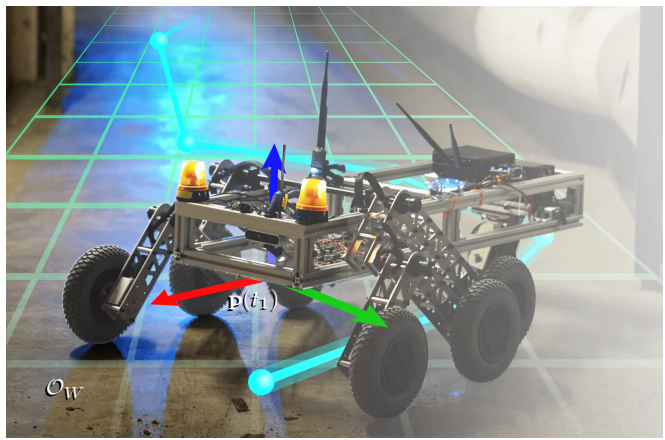

For visual sensing, 
our approach maintains a low 
sensory footprint with low-cost components: 
an RGB depth (RGB-D) camera to sense the environment. 
Most approaches tackling autonomous exploration use costly 
equipment such as 3D LiDARs~\cite{
kulkarni2022autonomous,tabib2022autonomous,tranzatto2022cerberus,roucek2020darpa,ebadi2020lamp,tardioli2019ground,dang2019graph,batinovic2021multi} and laser range finders~\cite{kim2022autonomous,surmann2003autonomous} instead. 
Even though approaches that utilize cheaper sensors, such as RGB-D cameras~\cite{
bircher2016receding,
dai2020fast,betz2022autonomous
}, RGB cameras~\cite{dang2019graph}, sonars~\cite{zhou2021smartphone,muller2021openbot}, and 2D LiDARs are studied~\cite{srinivasa2019mushr}, they often operate along 
more expensive 
hardware~\cite{dang2019graph
} or indoors only~\cite{bircher2016receding
} and have limited autonomy~\cite{
dai2020fast} or obstacle avoidance features~\cite{zhou2021smartphone,muller2021openbot}.
%

From a software perspective, 
recent efforts tackle autonomous exploration with prior learning~\cite{shrestha2019learned} or run on multiple robots~\cite{kulkarni2022autonomous,tranzatto2022cerberus,roucek2020darpa}, whereas 
approaches 
that require fewer computing resources are scarce~\cite{bircher2016receding,batinovic2021multi,faisal2021low,muller2021openbot}. 
Although some less computationally demanding approaches, such as those based on frontiers~\cite{kim2022autonomous,roucek2020darpa,batinovic2021multi}, graphs~\cite{kulkarni2022autonomous,tranzatto2022cerberus,dang2019graph}, grids~\cite{corah2019communication,tabib2022autonomous}, and random trees are studied, mixed approaches (i.e., a combination of the previous approaches) are preferred~\cite{shrestha2019learned,bircher2016receding,surmann2003autonomous,qiao2019sampling,dai2020fast}. In the presence of diverse sensing modalities, e.g., involving raw sensory data, topologies, semantics, etc., and outdoor environments~\cite{placed2022survey,batinovic2021multi}, mixed approaches 
maximize performance and resources~\cite{placed2022survey,bircher2016receding}. 
%
%
Our methodology is 
a mixed approach: a frontier- and sampling-based method that extends exploration 
with a path-following vector field 
from the aerial robotics domain~\cite{seewald2022energy,garcia2017guidance,seewaldphdthesis}. It
exploits the scarcity of resources while still running 
with good autonomy and obstacle avoidance features. 
Furthermore, our 
approach derives the 
robot's 
position using a state-of-the-art 
simultaneous localization and mapping (SLAM) 
algorithm~\cite{labbe2019rtab} and 
can operate in both unknown and GPS-denied environments. 
This allows the robot to explore its surroundings for longer and at lower update frequencies, 
utilizing cheaper computing~hardware. 

Utilizing these components 
with the open-source robot operating system (ROS) middleware, we additionally build  
a low-cost 
robotic platform -- RB5 in Figure~\ref{fig:0}, a wheeled mobile robot with rocker-bogie suspension -- capable of exploring indoor and outdoor environments autonomously. 
Comparable platforms in the literature comprise two degrees of freedom suspension with pivots~\cite{setterfield2013terrain,
faisal2021low} and provide rough terrain static adaptability. 
They are cheaper than, e.g., legged robots in terms of the price of sensors 
and operation, as they 
avoid obstacles without costly computations for gait adaptation and planning~\cite{muller2021openbot}.
The approach is generic in terms of portability to other mobile robots with price and computational constraints, but we target use cases where cheaper robotic explorers are preferred. These include nature conservation and surveying efforts~\cite{kirchgeorg2022multimodal} and education~\cite{betz2022autonomous,amster2020turtlebot}. 
The open-source software stack to replicate our approach is made available on the project repository webpage\footnoteref{link}.

The main \textbf{contributions} of this systems paper are 
\begin{enumerate*}[label={(\roman*)},font={\textit}]
  \item the \textit{implementation and design of a low-cost robot for autonomous long-term exploration} and 
  \item its \textit{feasibility and limitations analysis}. 
\end{enumerate*}
We demonstrate the exploration performance and obstacle avoidance features 
with a set of indoor and outdoor 
experiments in Section~\ref{sec:fe} and discuss the limitations of our low-cost exploration in Sec.~\ref{sec:lim}. The remainder of the paper is then structured as follows: Sec.~\ref{sec:pf} formalizes the problem of autonomous exploration, Sec.~\ref{sec:m} describes the approach from the software and hardware standpoints, and Sec.~\ref{sec:cf} concludes and provides future perspectives.


\section{Problem Description}
\label{sec:pf}
\noindent
The problem considered in this work 
is that of exploring an unknown bounded space, i.e., visiting 
every point 
within.
The robot is free to move except for 
some possible obstacles.
Formally, the problem 
is that of exploring a bounded volume $\mathcal{Q}\subseteq\mathbb{R}^3$ with respect to an inertial navigation frame $\mathcal{O}_W$. If the notation $[n]$ denotes a set of positive naturals up to $n\in\mathbb{N}_{>0}$ and $[n]_{>0}$ of strictly positive naturals, we are interested in collision-free trajectories that explore $\mathcal{Q}$ and avoid $n$ obstacles $\mathcal{Q}^{O_i}\subset\mathbb{R}^3,i\in[n]_{>0}$. We can approximate the space that delimits $\mathcal{Q}$ and $\mathcal{Q}^{O_i}$ for each $i$ with a set of vertices within which the two sets are contained.

\begin{pb}[Exploration]
  Consider sets of vertices $V:=\{\mathbf{v}_1,$ $\mathbf{v}_2,\dots\}$, $O_i:=\{\mathbf{o}_{i,1},\mathbf{o}_{i,2},\dots\}$ with $\mathbf{v}_j,\mathbf{o}_{i,k}\in\mathbb{R}^2,$ $\forall j\in[|V|],\,k\in[|O_i|]$ points w.r.t. $\mathcal{O}_W$. Let $V$ enclose $\mathcal{Q}$, $O_i$ $\mathcal{Q}^{O_i}$ per each $i\in[n]_{>0}$. The \textit{exploration problem} is the problem of finding the coverage that visits 
  every point $\mathbf{p}\in\mathcal{Q}\cap\mathcal{Q}^{O_1}\cap\mathcal{Q}^{O_2}\cap\cdots\cap\mathcal{Q}^{O_n}=:\mathcal{Q}^V$.
\end{pb}

Here the notation $|\cdot|$ denotes the cardinality and $\mathbb{R}$, $\mathbb{Z}$ are reals and integers. Bold notation is used for vectors.

Let $\phi$ be a path function, i.e., a function the robot tracks as it explores its surroundings in $\mathcal{Q}^V$, avoiding 
obstacles $\mathcal{Q}^{O_i}$.

\begin{defn}[Path function]\label{def:pf}
  $\phi:\mathbb{R}^2\rightarrow\mathbb{R}$ is a two-dimensional continuous and differentiable \textit{path function} of the $x$, $y$ components of $\mathbf{p}$.
\end{defn}

\begin{defn}[Coverage]\label{def:co}
  Given a tuple with a path function and its time component, $\langle\phi,t\rangle$, the \textit{coverage} is the collection of multiple tuples.
\end{defn}

The 
exploration approach 
(see Sec.~\ref{sec:le}) derives $\phi$ at each time step and adds it to the global ``coverage stack.'' The process ends once $\mathcal{Q}^V$ is covered.

\section{Systems Approach}
\label{sec:m}
\noindent
In this section, we detail the implementation and design choices in terms of 
our software and low-cost hardware for autonomous long-term 
exploration in Sec.~\ref{sec:le}~and~\ref{sec:md} respectively.

\subsection{Autonomous 
exploration}
\label{sec:le}

\begin{figure*}
  \begin{subfigure}[m]{0.33\textwidth}
    \centering
\begingroup%
  \makeatletter%
  \providecommand\color[2][]{%
    \errmessage{(Inkscape) Color is used for the text in Inkscape, but the package 'color.sty' is not loaded}%
    \renewcommand\color[2][]{}%
  }%
  \providecommand\transparent[1]{%
    \errmessage{(Inkscape) Transparency is used (non-zero) for the text in Inkscape, but the package 'transparent.sty' is not loaded}%
    \renewcommand\transparent[1]{}%
  }%
  \providecommand\rotatebox[2]{#2}%
  \newcommand*\fsize{\dimexpr\f@size pt\relax}%
  \newcommand*\lineheight[1]{\fontsize{\fsize}{#1\fsize}\selectfont}%
  \ifx\svgwidth\undefined%
    \setlength{\unitlength}{165bp}%
    \ifx\svgscale\undefined%
      \relax%
    \else%
      \setlength{\unitlength}{\unitlength * \real{\svgscale}}%
    \fi%
  \else%
    \setlength{\unitlength}{\svgwidth}%
  \fi%
  \global\let\svgwidth\undefined%
  \global\let\svgscale\undefined%
  \makeatother%
  \begin{picture}(1,0.75008871)%
    \lineheight{1}%
    \setlength\tabcolsep{0pt}%
    \put(0,0){\includegraphics[width=\unitlength,page=1]{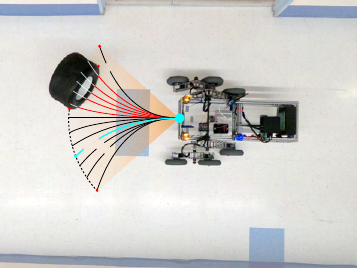}}%
    \put(0.24056443,0.33888954){\makebox(0,0)[lt]{\lineheight{1.25}\smash{\begin{tabular}[t]{l}\scriptsize $\phi_{t_0}$\end{tabular}}}}%
    \put(0.37953142,0.35465739){\makebox(0,0)[lt]{\lineheight{1.25}\smash{\begin{tabular}[t]{l}\scriptsize $\mathbf{p}({t_0})$\end{tabular}}}}%
    \put(0.2747604,0.18817172){\makebox(0,0)[lt]{\lineheight{1.25}\smash{\begin{tabular}[t]{l}\scriptsize $\mathbf{v}_2$\end{tabular}}}}%
    \put(0.28301384,0.62522182){\makebox(0,0)[lt]{\lineheight{1.25}\smash{\begin{tabular}[t]{l}\scriptsize $\mathbf{v}_1$\end{tabular}}}}%
    \put(0.28541002,0.54960308){\makebox(0,0)[lt]{\lineheight{1.25}\smash{\begin{tabular}[t]{l}\scriptsize $\mathbf{o}_{1,1}$\end{tabular}}}}%
    \put(0.08907668,0.42900242){\makebox(0,0)[lt]{\lineheight{1.25}\smash{\begin{tabular}[t]{l}\scriptsize $\mathbf{o}_{1,2}$\end{tabular}}}}%
  \end{picture}%
\endgroup%

    \caption{Initial detection of an obstacle wheel with $\phi_{t_0}$ selected so that it avoids the obstacle.}
    \label{fig:3-1}
  \end{subfigure}
  \hfill
  \begin{subfigure}[m]{0.32\textwidth}
    \centering
    \hspace*{-.15cm}
\begingroup%
  \makeatletter%
  \providecommand\color[2][]{%
    \errmessage{(Inkscape) Color is used for the text in Inkscape, but the package 'color.sty' is not loaded}%
    \renewcommand\color[2][]{}%
  }%
  \providecommand\transparent[1]{%
    \errmessage{(Inkscape) Transparency is used (non-zero) for the text in Inkscape, but the package 'transparent.sty' is not loaded}%
    \renewcommand\transparent[1]{}%
  }%
  \providecommand\rotatebox[2]{#2}%
  \newcommand*\fsize{\dimexpr\f@size pt\relax}%
  \newcommand*\lineheight[1]{\fontsize{\fsize}{#1\fsize}\selectfont}%
  \ifx\svgwidth\undefined%
    \setlength{\unitlength}{165bp}%
    \ifx\svgscale\undefined%
      \relax%
    \else%
      \setlength{\unitlength}{\unitlength * \real{\svgscale}}%
    \fi%
  \else%
    \setlength{\unitlength}{\svgwidth}%
  \fi%
  \global\let\svgwidth\undefined%
  \global\let\svgscale\undefined%
  \makeatother%
  \begin{picture}(1,0.75017746)%
    \lineheight{1}%
    \setlength\tabcolsep{0pt}%
    \put(0,0){\includegraphics[width=\unitlength,page=1]{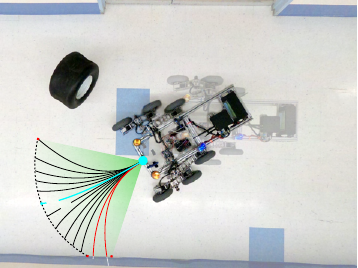}}%
    \put(0.32395125,0.04739132){\makebox(0,0)[lt]{\lineheight{1.25}\smash{\begin{tabular}[t]{l}\scriptsize $\mathbf{o}_{1,1}$\end{tabular}}}}%
    \put(0.0529328,0.37143693){\makebox(0,0)[lt]{\lineheight{1.25}\smash{\begin{tabular}[t]{l}\scriptsize $\mathbf{v}_1$\end{tabular}}}}%
    \put(0.26033995,0.30242335){\makebox(0,0)[lt]{\lineheight{1.25}\smash{\begin{tabular}[t]{l}\scriptsize $\mathbf{p}(t_1)$\end{tabular}}}}%
    \put(0.11012867,0.04239132){\makebox(0,0)[lt]{\lineheight{1.25}\smash{\begin{tabular}[t]{l}\scriptsize $\mathbf{o}_{1,2}$\end{tabular}}}}%
    \put(0.13310307,0.176507){\makebox(0,0)[lt]{\lineheight{1.25}\smash{\begin{tabular}[t]{l}\scriptsize $\phi_{t_1}$\end{tabular}}}}%
  \end{picture}%
\endgroup%

    \caption{The robot continues to track $\phi_{t_0}$ up to the next iteration. Here it finds a new trajectory $\phi_{t_1}$.}
    \label{fig:3-2}
  \end{subfigure}
  \begin{subfigure}[m]{0.33\textwidth}
    \centering
    \hspace*{.05cm}
    \input{figures/_GOPR0990.pdf_tex}
    \caption{The process continues up to when the entire space is explored.}
    \label{fig:3-3}
  \end{subfigure}
  \caption[Detail of our autonomous exploration methodology]{\textbf{Detail of our autonomous exploration methodology}. The 
  approach consists of the robot sampling the environment and searching for obstacles and unexplored areas. The 
  approach clusters the two groups into vertex sets and builds candidate path functions. From these, it selects the 
  trajectory w.r.t. a given cost function and iterates the operation at each step. In between the iterations, it tracks the trajectory, saving computational and sensing resources.}
  \label{fig:3}
\end{figure*}
\noindent
Our software stack 
consists of a mixed approach that combines frontier- and sampling-based methods. 
Here, with frontiers, we indicate 
boundaries between known and unknown space~\cite{
placed2022survey,dang2019graph}.
%

Our 
approach evaluates local frontiers at each step, samples the environment, and determines feasible candidate path functions $\phi$ that intersect $\mathcal{Q}^V$ (see Definition~\ref{def:pf}).
The next $\phi$ is selected so that the frontier is largest, but other cost functions are possible (see Sec.~\ref{sec:cf}). The collection of the candidate path functions forms the coverage (see Def.~\ref{def:co}).
The 
approach then derives a path-following vector field that points to $\phi$ at any point and guides the robot utilizing the gradient descent algorithm. This allows the robot to, e.g., follow the coverage path for longer and in real-time compared to approaches that utilize frontiers only, decreasing computational requirements (see Sec.~\ref{sec:fe}).

To derive the path-following vector field, let the gradient of $\phi$ be defined
\begin{equation}
  \nabla\phi:=\begin{bmatrix}
    \partial\phi
    /\partial\mathbf{p}_x\\
    \partial\phi
    /\partial\mathbf{p}_y
  \end{bmatrix},
\end{equation}
where $\partial\phi/\partial\mathbf{p}$ is the partial differential, and $\mathbf{p}_x$, $\mathbf{p}_y$ are the $x$ and $y$ components of $\mathbf{p}$.
The path-following vector field points in the direction where $\phi$ maximally locally increases. To assign the direction to each point, we use the construct of vector fields, which is common in other motion planning literature~\cite{
garcia2017guidance,goncalves2010vector}
\begin{equation}\label{eq:vecf}
  \Phi(t,\phi):={\textstyle \bigcup\limits_{\mathbf{p}(t)\in\mathcal{Q}}}\nabla\phi(\mathbf{p}(t)).
\end{equation}

We modify the vector field in Equation~(\ref{eq:vecf}) to point to the contour of the path function $\phi$ rather than its local maxima, as proposed in~\cite{garcia2017guidance}
\begin{equation}\label{eq:pfvf}
  \Delta\phi(\mathbf{p}(t)):=E\nabla\phi(\mathbf{p}(t))-k_e\phi(\mathbf{p}(t))\nabla\phi(\mathbf{p}(t)),
\end{equation}
where $E\nabla\phi$ points perpendicularly to the gradient and $\phi\nabla\phi$ to $\phi$ at $k_e\in\mathbb{R}_{>0}$ rate. $E$ is the following direction, i.e.,
\begin{equation}
  E=\begin{bmatrix}
    1 & 0\\ 0 & -1
  \end{bmatrix},
\end{equation}
is counterclockwise and $-E$ clockwise directions.

Let the path-following equivalent of Eq.~(\ref{eq:vecf}) be 
\begin{equation}
  \Phi_\phi(t,\phi):={\textstyle \bigcup\limits_{\mathbf{p}(t)\in\mathcal{Q}}}\Delta\phi(\mathbf{p}(t)).
\end{equation}

\begin{algorithm}[t]
  \begin{algorithmic}[1]
    \small
    \FORALL{$t\in\mathcal{T}$}
      \STATE \textbf{if} $\mathcal{P}\cap\mathcal{Q}=\varnothing$ \textbf{then return }$\langle\phi,t\rangle$\vspace*{.3ex}\label{alg:cs}
      \STATE $\mathcal{Q}^V_t:=\{O_{1,t},O_{2,t},\dots,O_{n,t},V_t\}\gets$ sensor readings\label{alg:vd}\vspace*{.3ex}
      \IF{$\mathcal{Q}_t^V\neq\mathcal{Q}_{t-1}^V$}\vspace*{.3ex}
        \STATE $\{\phi_{1,t},\phi_{2,t},\dots\}\gets$ $\phi$s in Def.~\ref{def:pf}, inters. $\mathcal{Q}^V\cap \Psi(\mathcal{Q}^V_t)$\vspace*{-1.6ex}
        \STATE \textbf{if} $\phi_t:=\{\phi_{1,t},\phi_{2,t},\dots\}=\varnothing$ \textbf{then }the robot is stuck\label{alg:mpty}\vspace*{.3ex}
        \STATE \textbf{else}
        \STATE $\,\,\,\,\,\,\phi_t\gets \argmax_{\phi}l(\phi_t,t,\mathcal{Q}_t^V)$ in Eq.~(\ref{eq:cost})\label{alg:am}\vspace*{.3ex}
        \STATE $\,\,\,\,\,\,\langle\phi,t\rangle\gets\langle\phi,t\rangle\cup\langle\phi_t,t\rangle$ in Def.~\ref{def:co}\vspace*{.3ex}
        \STATE $\,\,\,\,\,\,\mathcal{P}\gets\mathcal{P}\cup \Psi(\mathcal{Q}_t^V)$\label{alg:vp}
        \STATE \textbf{end if}
        \vspace*{.3ex}
      \ENDIF
      \STATE $\varphi(t,\mathbf{p}(t))\gets\varphi(t-1,\mathbf{p}(t-1))+\theta\Delta\phi(\mathbf{p}(t))$ in Eq.~(\ref{eq:pfvf})\vspace*{.3ex}\label{alg:vf}
    \ENDFOR
  \end{algorithmic}
  \caption{Derivation of the exploration coverage $\langle\phi,t\rangle$}\label{alg}
\end{algorithm}

The path-following vector field is summarized in the pseudo-code in Algorithm~\ref{alg}, with the gradient descent in Line~\ref{alg:vf}. The vector $\varphi\in\mathbb{R}^2$ points the robot in the direction of the path function $\phi$ with a scalar step size $\theta\in\mathbb{R}_{>0}$. The algorithm runs at each time step in $\mathcal{T}$, 
but practically there might be different time steps at different times (see Sec.~\ref{sec:fe}). 
In Line~\ref{alg:cs}, the algorithm evaluates if the bounded volume $\mathcal{Q}$ is covered utilizing $\mathcal{P}\subseteq\mathbb{R}^3$ updated in Line~\ref{alg:vp}, where the function $\Psi:\mathbb{R}^{2n}\times\mathbb{R}^2\rightarrow\mathbb{R}^{3n}\times\mathbb{R}^3$ maps the vertices to the volume. The vertices of the local free space $\mathcal{Q}^V_t$ in Line~\ref{alg:vd} are derived from sensor readings, assuming the presence of an RGB-D camera. The 
approach reads the camera's point cloud, clustering the obstacles $O_{1,t},O_{2,t},\dots$ by checking if the distance between consecutive points in space is within a given threshold $\varepsilon\in\mathbb{R}_{>0}$. The vertices of the free space at time instant $t$, $V_t$ are simply the limits of the sensor's field of view.

The remaining lines 
compute the feasible path functions $\{\phi_{1,t},\phi_{2,t},\dots\}$ by intersecting the local free space $\Psi(\mathcal{Q}^V_t)$ with possible candidate trajectories that have their final points laying at the edges of $\mathcal{Q}^V_t$, i.e., splines of the form
\begin{equation}\label{eq:spln}
  a(x-\mathbf{p}_x)^3+b(x-\mathbf{p}_x)^2+c(x-\mathbf{p}_x)+d-y=0,
\end{equation}
where $a,b,c\in\mathbb{R}$ are the coefficients of the spline. 
The best trajectory is then derived via the cost function $l$ in Line~\ref{alg:am}, utilizing the intersection of the largest frontier. Formally
\begin{equation}\label{eq:cost}\begin{split} 
  l:=\bigl\{\lVert \mathbf{p}_1-\mathbf{p}_2\rVert\,|\,&\exists\,\mathbf{p}_1,\mathbf{p}_2\in\Psi(\mathcal{Q}_t^V),i\in[|\phi_t|]\\
  &\text{ s.t. }\mathbf{p}_1\neq\mathbf{p}_2,\phi_{i,t}(\mathbf{p}_1-\mathbf{p}_2)\trianglelefteq 0\bigr\},
\end{split}\end{equation}
where the operator $\trianglelefteq$ evaluates $\phi$ on a given $\varepsilon\in\mathbb{R}_{>0}$, i.e., $|\phi_{i,t}(\mathbf{p}_1-\mathbf{p}_2)|\leq\varepsilon$ and in such a way that the middle path functions of the largest subset of the contiguous path functions are selected preferably, e.g., if the largest subset is $\{\phi_{1,t},\phi_{2,t},\dots,\phi_{5,t}\}$, $\phi_{3,t}$ is selected.
In this way, if there are no obstacles, Eqs.~(\ref{eq:spln}--\ref{eq:cost}) are such that $\phi$ is a line parallel to the direction of travel. 

The algorithm is illustrated in Fig.~\ref{fig:3}. At each iteration, the robot samples the environment and derives a set of 
candidate path functions $\{\phi_{1,t},\phi_{2,t},\dots\}$. If there is no obstacle ahead, the optimal function per iteration $\phi_t$ is a line parallel to the robot's direction of travel (see Fig.~\ref{fig:3-3}). If there are obstacles, the 
approach selects the trajectory via the cost function $l$, $\phi_t$, which goes through the middle of the largest frontier (see Fig.~\ref{fig:3-1}~and~\ref{fig:3-2} for 
obstacles ``wheel'' and ``wall'').

The 
algorithm provides a way to explore space $\mathcal{Q}$ and avoid obstacles $\mathcal{Q}^{O_i}$. 
Nevertheless, there are configurations at which there are no feasible trajectories, i.e., if $\{\phi_{1,t},\phi_{2,t},\dots\}=\varnothing$ in Line~\ref{alg:mpty}. In this scenario, 
our approach allows a human to intervene via 
wireless or LoRa communication technology. The robot can be then teleoperated over long distances -- studies from the internet-of-things domain~\cite{shanmuga2020survey
} report a range of up to five kilometers in an urban setting -- and with 
inexpensive hardware equipment (two LoRa bundles). The 
approach we propose utilizes a web interface to parse human commands into our custom communication protocol, which utilizes the LoRa physical layer's payload to transfer $\varphi$'s $x$ and $y$ components (see Line~\ref{alg:vf}). The current setup is limited to control and position commands.

To derive a map of the environment and to keep 
track of the robot within -- in Line~\ref{alg:vf} -- our 
approach uses a state-of-the-art 
SLAM algorithm from the literature~\cite{labbe2019rtab}. The robot's location is also used to determine whether the exploration is complete in Line~\ref{alg:cs}. 
An earlier iteration of the work exploited a different SLAM algorithm from the visual SLAM community~\cite{campos2021orb}, showing that some of the 
components are interchangeable.

The 
software stack 
is distributed under the popular open-source CC BY-NC-SA license\footnoteref{link}. 
It is composed of three distinct components. 
\begin{enumerate*}[label={(\roman*)},font={\textit}]
  \item A ``ground robot'' ROS2 
package implements the communication with a base station using either the IEEE 802.11 wireless communication or long-range LoRa technology. The package further 
contains serial communication with the microcontroller implemented in Arduino and the vertices detection (see Algorithm~\ref{alg}). 
  \item A ``ground navigation'' ROS package collects point clouds from an RGB-D camera 
and other data from the SLAM algorithm~\cite{labbe2019rtab} and ports them into ROS2. Finally, 
  \item a ``base server'' implements the necessary functionality for remote human intervention.
\end{enumerate*}
Both ground robot and ground navigation are implemented in C++ in ROS2 and ROS respectively, whereas base server is in PHP and JavaScript.

\subsection{Low-cost 
hardware design}
\label{sec:md}
\noindent
Our RB5 experimental robotic platform 
adopts a rocker-bogie suspension system~\cite{bickler1989articulated} found on NASA's rovers including Sojourner and Curiosity, which has compelling tradeoffs in terms of autonomy and obstacle avoidance (see Sec.~\ref{sec:fe}). On either side of the robot, an upside-down V-shaped linkage called the rocker pivots about an axis on the robot frame. The rocker has a wheel at one end and a smaller V-shaped linkage on the other arm. The smaller linkage, called the bogie, can pivot about an axis on the rocker and has two wheels at its tips. The articulated nature of the rocker-bogie suspension allows the mobile robot to adapt to uneven terrains~\cite{
faisal2021low} as the rocker and bogie pivot to maintain wheel contact. 
Each of the six wheels in the rocker-bogie suspension is actuated by a DC gear motor, whereas the rotational degrees of freedom in the rocker-bogie suspensions are passive. Since the wheels are all parallel and cannot rotate out of the plane, the robot uses the same actuation strategy as that of a differential drive vehicle to move straight and make turns by controlling the left and right sets of wheels in the same and opposite directions. Given that RB5 has multiple wheels on each side, its ability to make turns is reduced compared to a differential drive vehicle. Due to its extended body length, RB5 incorporates a caster wheel in the back to support the rear end of the frame.

\begin{figure}[t]
  \vspace*{-.25cm}
  \begin{minipage}[t]{0.58\columnwidth}
    \hspace*{-.5cm}
    \input{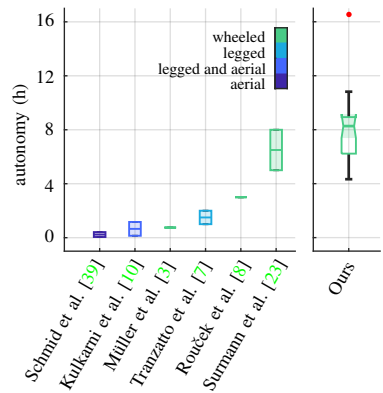}
  \end{minipage}\hfill
  \begin{minipage}[t]{0.41\columnwidth}
    \vspace*{-5.25cm}
    \centering
    \caption[Autonomy for different classes of mobile robots]{
    \textbf{Autonomy for different classes of mobile robots}. 
    Autonomy is reported in hours between the time the battery is fully charged to discharged for our RB5 explorer and other approaches that report the value in the literature. Even though the metric is use case- and battery-dependent, the data show that the reported autonomy for wheeled robots is higher than the reported autonomy for legged, combination of legged and aerial, and aerial~robots.
    }
    \label{fig}
  \end{minipage}
  \vspace*{-1cm}
\end{figure}

The robot frame's dimensions are 914 by 330 millimeters, and the robot's bounding box dimensions are 991 by 762 mm. The frame consists of one-inch aluminum extrusions and acrylic sheets, and the rocker and bogie linkages are assembled from aluminum sheets and standoffs. The pivots of the bogie and rocker sit at 240 and 330 mm from the ground respectively, providing a clearance of approximately 190 mm beneath the frame, i.e., RB5 can clear obstacles passively up to 19 cm. The two wheels on each bogie linkage are coplanar, but the wheel on the corresponding rocker linkage is closer to the medial plane of the robot. Motor control is performed by a Teensy (R) 4.0 microcontroller board sending PWM commands to six DRV8871 motor driver boards. An onboard 24 volts LiFePO${}_\text{4}$ battery provides power for the microcontroller, motor drives, and computing hardware.

\begin{figure*}[b]
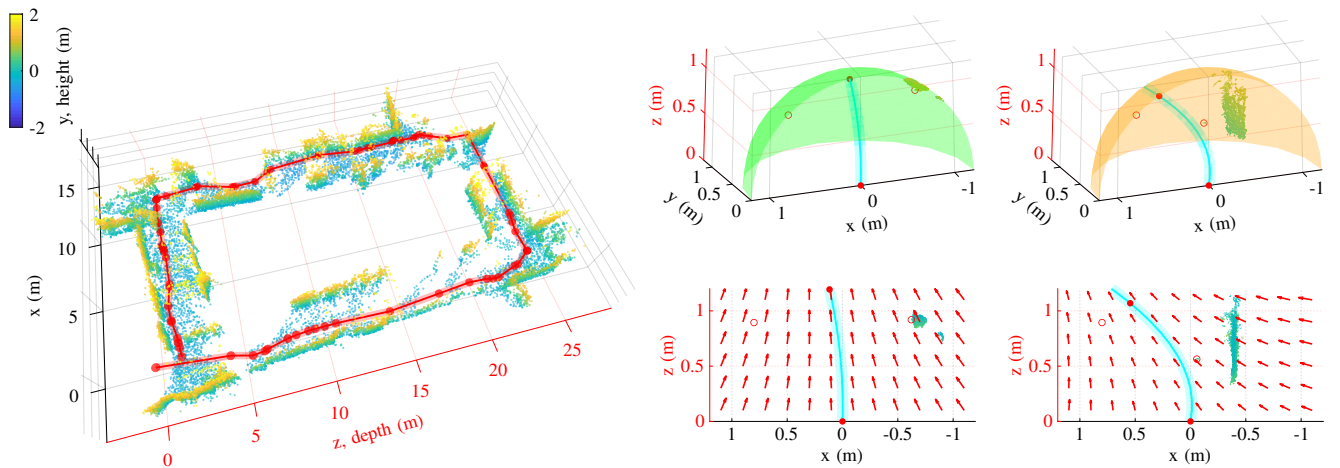

  \vspace*{-.7cm}
  \begin{subfigure}[m]{0.48\textwidth}
    \centering
    \hspace*{-1.05cm}
    \input{figures/map_plot_2023_02_16.pdf_tex}
    \vspace*{-.655cm}
    \caption{Point cloud view of a structured indoor environment with visible countors of the exploration space. Points are colored for different heights.}
    \label{fig:1-3}
  \end{subfigure}
  \hfill
  \begin{subfigure}[m]{0.25\textwidth}
    \centering
    \input{figures/nav_plot_2023_02_16.pdf_tex}
    \caption{The first detection of an obstacle door. A path function is selected to avoid the obstacle.}
    \vspace*{-.7cm}
    \label{fig:1-1}
  \end{subfigure}
  \hfill
  \begin{subfigure}[m]{0.25\textwidth}
    \centering
    \input{figures/_nav_plot_2023_02_16.pdf_tex}
    \caption{The new path function is selected at the next time step as the obstacle occurrence is observed closer.}
    \vspace*{-.7cm}
    \label{fig:1-2}
  \end{subfigure}
  \caption[Results for a structured environment]{\textbf{Results for a structured environment}. Experimental results are reported for a structured indoor environment, a university hallway composed of four connected corridors for a total length of approximately 80 meters. The view includes the point cloud in Fig.~\ref{fig:1-3} and the detail of the algorithm for obstacle avoidance and detection at successive time steps in Fig.~\ref{fig:1-1}~and~\ref{fig:1-2}. The points in the point cloud are filtered to report one point every 250. The colors of the spheres in Figs.~\ref{fig:1-1}--\ref{fig:1-2} indicate the proximity of an obstacle (orange indicates close proximity) and arrows the path-following vector field in Eq.~(\ref{eq:pfvf}). The robot's trajectory is in red and red dots indicate SLAM's registration points.}
  \label{fig:1}
\end{figure*}
\begin{figure*}[t]
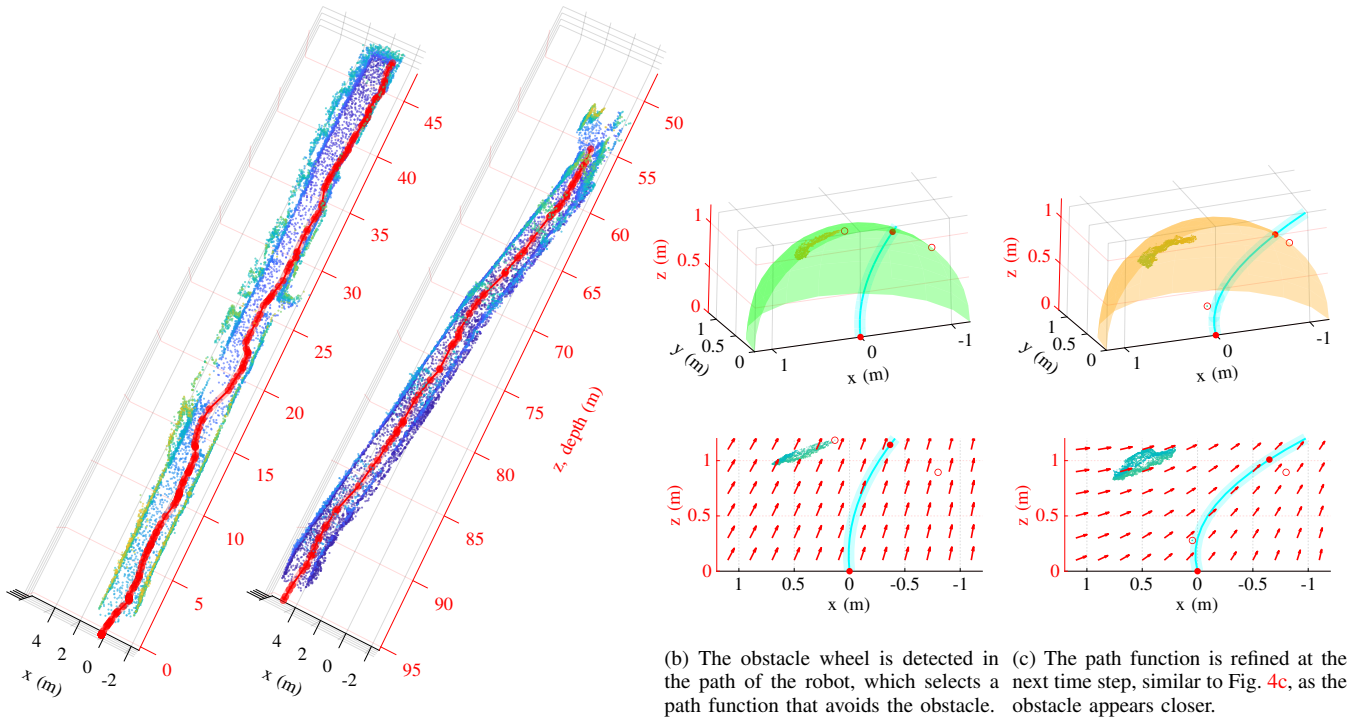

  \vspace*{-.3cm}
  \begin{subfigure}[m]{0.48\textwidth}
    \centering
    \hspace*{-.9cm}
    \input{figures/map_plot_2023_01_26_and_02_08.pdf_tex}
    \vspace*{-.5cm}
    \caption{Point cloud view of an unstructured indoor environment (left) and an underground tunnel (right) with visible contours of the exploration space. The color scale is the same as in Fig.~\ref{fig:1-3}.}
    \label{fig:2-1}
  \end{subfigure}
  \hfill
  \begin{subfigure}[m]{0.25\textwidth}
    \centering
    \input{figures/nav_plot_2023_01_26_and_02_08_1.pdf_tex}
    \caption{The obstacle wheel is detected in the path of the robot, which selects a path function that avoids the obstacle.}
    \vspace*{-1cm}
    \label{fig:2-2}
  \end{subfigure}
  \hfill
  \begin{subfigure}[m]{0.25\textwidth}
    \centering
    \input{figures/_nav_plot_2023_01_26_and_02_08_1.pdf_tex}
    \caption{The path function is refined at the next time step, similar to Fig.~\ref{fig:1-2}, as the obstacle appears closer.}
    \vspace*{-1cm}
    \label{fig:2-3}
  \end{subfigure}
  \caption[Results for an unstructured environment]{\textbf{Results for an unstructured environment}. Experimental results are reported for an unstructured indoor environment and an underground tunnel for a total length of approximately 100 meters. The view of the point cloud in Fig.~\ref{fig:2-1} is filtered to report one point every 500. The detail of the algorithm for successive time steps is shown in Figs.~\ref{fig:2-2}--\ref{fig:2-3}, similar to Fig.~\ref{fig:1}.}
  \label{fig:2}
  \vspace*{-.25cm}
\end{figure*}

\section{Experiments}
\label{sec:fe}
\noindent
In order to demonstrate 
our approach, we conduct a set of 
field experiments 
involving our RB5 experimental robotic platform 
in a variety of environments, including indoor structured, unstructured underground, and outdoors. In each, the microcontroller executes a finite set of motion primitives via velocity control. These primitives are transmitted serially to the microcontroller 
from RB5's onboard computing hardware, an NVIDIA (R) Jetson Xavier NX (TM) embedded board, which runs our autonomous 
exploration software stack. 
The computing hardware mounts peripherals for visual sensing and for communication. The former group consists of 
an Intel (R) RealSense (TM) D435 RGB-D camera, and the latter 
consists 
of a LoRa wireless network bundle with the RN2903 module and an Intel~(R) AX200 network card for standard wireless communication via 802.11 protocol when, e.g., RB5 is in reach of an available wireless network.
All the software components in charge of the exploration 
(detailed at the end of Sec.~\ref{sec:le}) run in real-time onboard RB5. Additional processing is possible, e.g., 
via the ROS network.

Fig.~\ref{fig} compares our hardware approach to others. 
``Autonomy,'' which is related to instantaneous energy consumption, is reported in hours between the time the battery is fully charged to discharged -- the time when the robot can actively explore its surroundings -- and is compared to representative approaches in the literature tackling autonomous exploration. For our RB5, the minimum and maximum are approximately four hours and twenty minutes and sixteen and a half hours when the robot respectively moves at full speed and is not moving (red outlier). The first quartile is six hours and ten minutes, the third is nine hours. The median is eight hours and twenty minutes when the average velocity is two-thirds of the maximum. Even though the metric is often use case- and battery-dependent (i.e., in \cite{roucek2020darpa,tranzatto2022cerberus} the objective is to explore the surroundings in the shortest time), the data show that the reported autonomy for wheeled robots~\cite{roucek2020darpa,surmann2003autonomous} is generally higher than the reported autonomy for legged~\cite{tranzatto2022cerberus}, combination of legged and aerial~\cite{kulkarni2022autonomous}, and aerial robots~\cite{schmid2020efficient}.
Here \cite{muller2021openbot} is an outlier as it uses a small wheeled robot.


Fig.~\ref{fig:1} shows experimental results for a structured indoor environment, a university hallway on the second floor of a multistory building. The hallway is composed of four connected corridors for a total approximate length of 80 meters in a closed circuit. 
The resulting point cloud is shown in Fig.~\ref{fig:1-3}, where the color scheme in the top-left indicates the different heights of points in the point cloud. 
%
%
Figs.~\ref{fig:1-1}--\ref{fig:1-2} show a detail of the algorithm in the experiment in terms of obstacle detection and avoidance. Here RB5 detects an obstacle, a ``door'' with a surrounding wall as it travels through the hallway at approximately 15 and 0 on the respective z- and x-axis. Fig.~\ref{fig:1-1} shows the initial detection of the obstacle on top. The vertices $V, O_i$ are the empty red circles and represent the field of view on the left and the edge of the obstacle on the right. On the bottom is the path-following vector field from Eq.~(\ref{eq:pfvf}) in red and the path function $\phi_t$ in cyan. Fig.~\ref{fig:1-2} shows the following time step when the robot has to perform a sharper maneuver to avoid the obstacle.

Fig.~\ref{fig:2} shows experimental results for an unstructured environment, a hallway connecting to an underground tunnel (in Fig.~\ref{fig:0}) on the respective left and right sides of Fig.~\ref{fig:2-1}. The hallway and tunnel combined have an approximate length of 100 meters. Conversely to the experiment in Fig.~\ref{fig:1}, this experiment showcases an open circuit, i.e., the exploration is concluded when a specific frontier is encountered. Figs.~\ref{fig:2-2}--\ref{fig:2-3} show the obstacle detection, similar to Figs.~\ref{fig:1-1}--\ref{fig:1-2}, for a wheel placed close to the left edge of the first length of the figure wide approximately 0.42 meters. The trajectory of the robot avoiding the obstacle is to be observed in Fig.~\ref{fig:2-1} between 15 and 20 meters on the z-axis. The figure also demonstrates the remote human intervention via LoRa, as the robot is stuck at the entrance of the underground tunnel (approximately 50 meters on the z-axis in Fig.~\ref{fig:2-1}).

The turning direction $E$ in Eq.~(\ref{eq:pfvf}) is positive for left turns (see Figs.~\ref{fig:1-1}--\ref{fig:1-2}) and negative for right turns (see Figs.~\ref{fig:2-3}--\ref{fig:2-2}). The turning rate $k_e$ is derived empirically similar to other literature~\cite{seewald2022energy,garcia2017guidance} and is 0.05, 0.1, and 0.4 depending on the turning maneuver, i.e., it is 0.05 when $\phi_t$ is a line (or close to it), 0.4 when a sharp curve in respectively Fig.~\ref{fig:1-3}~and~\ref{fig:2-2}, and 0.1 otherwise.  
The points in the point cloud are adjusted for height and length and filtered for visualization purposes, i.e., we have reported one point every 250, every 500, etc., in Fig.~\ref{fig:1-3}~and~\ref{fig:2-1}.

\section{Limitations of a Low-Cost Explorer}\label{sec:lim}
\noindent
In this section, we discuss 
the limitations of the current approach from both software and hardware perspectives.

Software-wise, a negative of the low-cost approach is a reduced density of the point cloud, as visualized in Fig.~\ref{fig:1}, where between, e.g., 15 and 20 meters on the z-axis and zero and five meters on the x-axis there are significantly fewer points in the point cloud than in other parts of the figure. 
The algorithm here keeps track (see Line~\ref{alg:vf}) of the path function $\phi_t$ (see Line~\ref{alg:am}) in the event of, e.g., the computing hardware being busy while executing other tasks such as communication. While specific to the computing hardware onboard RB5, the occurrence is expected with lower-performing computing hardware. It is due to the unpredictable nature of the execution, which is a common occurrence in the literature, especially if involving heterogeneous elements, i.e., CPU, GPU, and microcontrollers~\cite{seewald2019coarse}.
Despite a lower update frequency, the approach maintains its obstacle avoidance and navigation capabilities, with a nominal frequency of one to ten hertz.

Hardware-wise, a hurdle that we encountered 
is that many components are still 
expensive and limited in variety. Prior work has been 
opting for expensive servo motors or well-established electric motor manufacturers.
Furthermore, existing low-cost 
kits such as the TurtleBot~\cite{amster2020turtlebot} are limited to deployment in 
environments that are not physically demanding. There is still a large gap in low-cost robot hardware that can be tested in challenging conditions. Due to the lack of a common specification 
for a rough terrain environment, there are no performance or life cycle requirements to meet in the 
design process; therefore, it is difficult to develop a low-cost generalized mobile robot for rough terrains.

\section{Conclusion and Future Directions}
\label{sec:cf}
\noindent
This paper consists of an 
approach and an experimental robotic platform 
for low-cost autonomous long-term exploration in both indoor and outdoor 
environments. While comparable with other 
approaches tackling autonomous exploration, 
our approach 
operates in the presence of fewer sensory and computing requirements. Requiring only an RGB-D camera, all the exploration is computed in real-time on low-power computing hardware that is cheaper compared to the existing literature 
in similar settings.

The exploration consists of 
a 
mixed approach -- a frontier- and sampling-based method from the literature 
extended with a path-following vector field 
from the aerial robotics domain 
-- which allows the robot to operate at lower update frequencies. The position is from a state-of-the-art SLAM algorithm. 
Human intervention, if required, is implemented via a novel methodology based on the LoRa low-power long-range communication technology 
from the internet-of-things domain. 
Requiring only two low-cost LoRa bundles for communication, the approach enables operations on long distances with a custom communication protocol with no significant impact on price and resources as opposed to existing methodologies based on a mesh of devices. 

To enable further savings, we are currently extending the approach to account for energy requirements and to 
different cost functions in Eq.~(\ref{eq:cost}). Applicability to different use cases is also being investigated via, e.g., implementation of feature detection and tracking and extension of the custom LoRa communication protocol. 

{\small
\section*{Acknowledgments}
\noindent
We thank Paedyn~Gomes, Victoria~Ereskina, and Beau~Birdsall for their contribution to the mechanical design.

{\small\addtolength{\textheight}{-1.0cm}
\bibliographystyle{IEEEtran}
\bibliography{rb5-paper}

\begin{thebibliography}{10}
\providecommand{\url}[1]{#1}
\csname url@samestyle\endcsname
\providecommand{\newblock}{\relax}
\providecommand{\bibinfo}[2]{#2}
\providecommand{\BIBentrySTDinterwordspacing}{\spaceskip=0pt\relax}
\providecommand{\BIBentryALTinterwordstretchfactor}{4}
\providecommand{\BIBentryALTinterwordspacing}{\spaceskip=\fontdimen2\font plus
\BIBentryALTinterwordstretchfactor\fontdimen3\font minus
  \fontdimen4\font\relax}
\providecommand{\BIBforeignlanguage}[2]{{%
\expandafter\ifx\csname l@#1\endcsname\relax
\typeout{** WARNING: IEEEtran.bst: No hyphenation pattern has been}%
\typeout{** loaded for the language `#1'. Using the pattern for}%
\typeout{** the default language instead.}%
\else
\language=\csname l@#1\endcsname
\fi
#2}}
\providecommand{\BIBdecl}{\relax}
\BIBdecl

\bibitem{lluvia2021active}
I.~Lluvia, E.~Lazkano, and A.~Ansuategi, ``Active mapping and robot
  exploration: A survey,'' \emph{Sensors}, vol.~21, no.~7, p.~26, 2021.

\bibitem{placed2022survey}
J.~A. Placed, J.~Strader, H.~Carrillo \emph{et~al.}, ``A survey on active
  simultaneous localization and mapping: State of the art and new frontiers,''
  \emph{IEEE Transactions on Robotics}, vol.~39, no.~3, pp. 1686--1705, 2023.

\bibitem{muller2021openbot}
M.~M{\"u}ller and V.~Koltun, ``{OpenBot: T}urning smartphones into robots,'' in
  \emph{International Conference on Robotics and Automation (ICRA)}.\hskip 1em
  plus 0.5em minus 0.4em\relax IEEE, 2021, pp. 9305--9311.

\bibitem{zhou2021smartphone}
B.~Zhou, Z.~Wu, and X.~Liu, ``Smartphone-based robot indoor localization using
  inertial sensors, encoder and map matching,'' in \emph{International
  Conference on Automation, Control and Robots (ICACR)}.\hskip 1em plus 0.5em
  minus 0.4em\relax IEEE, 2021, pp. 145--149.

\bibitem{srinivasa2019mushr}
S.~Srinivasa, P.~Lancaster, J.~Michalove \emph{et~al.}, ``{MuSHR}: A low-cost,
  open-source robotic racecar for education and research,'' p.~4, 2019, arXiv
  preprint,
  {\tt\footnotesize\href{https://doi.org/10.48550/arXiv.1908.08031}{doi.org/10.48550/arXiv.1908.08031}}.

\bibitem{faisal2021low}
S.~M.~F. Faisal, T.~Rahman, and M.~A. Kabir, ``A low-cost rough terrain
  explorer robot fabrication using rocker bogie mechanism,'' in
  \emph{International Conference on Computer, Communication, Chemical,
  Materials and Electronic Engineering (IC4ME2)}.\hskip 1em plus 0.5em minus
  0.4em\relax IEEE, 2021, p.~4.

\bibitem{tranzatto2022cerberus}
M.~Tranzatto, F.~Mascarich, L.~Bernreiter \emph{et~al.}, ``{CERBERUS:
  A}utonomous legged and aerial robotic exploration in the tunnel and urban
  circuits of the {DARPA S}ubterranean {C}hallenge,'' \emph{Field Robotics},
  vol.~2, pp. 274--324, 2022.

\bibitem{roucek2020darpa}
T.~Rou{\v{c}}ek, M.~Pecka, P.~{\v{C}}{\'i}{\v{z}}ek \emph{et~al.}, ``{DARPA
  S}ubterranean {C}hallenge: {M}ulti-robotic exploration of underground
  environments,'' in \emph{International Conference on Modelling and Simulation
  for Autonomous Systems (MESAS)}.\hskip 1em plus 0.5em minus 0.4em\relax
  Springer, 2020, pp. 274--290.

\bibitem{tabib2022autonomous}
W.~Tabib, K.~Goel, J.~Yao \emph{et~al.}, ``Autonomous cave surveying with an
  aerial robot,'' \emph{IEEE Transactions on Robotics}, vol.~38, no.~2, pp.
  1016--1032, 2022.

\bibitem{kulkarni2022autonomous}
M.~Kulkarni, M.~Dharmadhikari, M.~Tranzatto \emph{et~al.}, ``Autonomous teamed
  exploration of subterranean environments using legged and aerial robots,'' in
  \emph{International Conference on Robotics and Automation (ICRA)}.\hskip 1em
  plus 0.5em minus 0.4em\relax IEEE, 2022, pp. 3306--3313.

\bibitem{ebadi2020lamp}
K.~Ebadi, Y.~Chang, M.~Palieri \emph{et~al.}, ``{LAMP: L}arge-scale autonomous
  mapping and positioning for exploration of perceptually-degraded subterranean
  environments,'' in \emph{International Conference on Robotics and Automation
  (ICRA)}.\hskip 1em plus 0.5em minus 0.4em\relax IEEE, 2020, pp. 80--86.

\bibitem{khairuldanial2019mobile}
K.~Ismail, R.~Liu, J.~Zheng \emph{et~al.}, ``Mobile robot localization based on
  low-cost {LTE} and odometry in {GPS}-denied outdoor environment,'' in
  \emph{International Conference on Robotics and Biomimetics (ROBIO)}.\hskip
  1em plus 0.5em minus 0.4em\relax IEEE, 2019, pp. 2338--2343.

\bibitem{voigtlander20175g}
F.~Voigtl{\"a}nder, A.~Ramadan, J.~Eichinger \emph{et~al.}, ``{5G} for
  robotics: Ultra-low latency control of distributed robotic systems,'' in
  \emph{International Symposium on Computer Science and Intelligent Controls
  (ISCSIC)}.\hskip 1em plus 0.5em minus 0.4em\relax IEEE, 2017, pp. 69--72.

\bibitem{delgado2022oros}
C.~Delgado, L.~Zanzi, X.~Li \emph{et~al.}, ``{OROS: O}rchestrating {ROS}-driven
  collaborative connected robots in mission-critical operations,'' in
  \emph{International Symposium on a World of Wireless, Mobile and Multimedia
  Networks (WoWMoM)}.\hskip 1em plus 0.5em minus 0.4em\relax IEEE, 2022, pp.
  147--156.

\bibitem{cadena2016past}
C.~Cadena, L.~Carlone, H.~Carrillo \emph{et~al.}, ``Past, present, and future
  of simultaneous localization and mapping: Toward the robust-perception age,''
  \emph{IEEE Transactions on Robotics}, vol.~32, no.~6, pp. 1309--1332, 2016.

\bibitem{eldemiry2022autonomous}
A.~Eldemiry, Y.~Zou, Y.~Li \emph{et~al.}, ``Autonomous exploration of unknown
  indoor environments for high-quality mapping using feature-based {RGB-D
  SLAM},'' \emph{Sensors}, vol.~22, no.~14, p.~16, 2022.

\bibitem{corah2019communication}
M.~Corah, C.~O'Meadhra, K.~Goel \emph{et~al.}, ``Communication-efficient
  planning and mapping for multi-robot exploration in large environments,''
  \emph{IEEE Robotics and Automation Letters}, vol.~4, no.~2, pp. 1715--1721,
  2019.

\bibitem{shanmuga2020survey}
J.~P. Shanmuga~Sundaram, W.~Du, and Z.~Zhao, ``A survey on {LoRa} networking:
  Research problems, current solutions, and open issues,'' \emph{IEEE
  Communications Surveys \& Tutorials}, vol.~22, no.~1, pp. 371--388, 2020.

\bibitem{tardioli2019ground}
D.~Tardioli, L.~Riazuelo, D.~Sicignano \emph{et~al.}, ``Ground robotics in
  tunnels: {K}eys and lessons learned after 10 years of research and
  experiments,'' \emph{Journal of Field Robotics}, vol.~36, no.~6, pp.
  1074--1101, 2019.

\bibitem{dang2019graph}
T.~Dang, F.~Mascarich, S.~Khattak \emph{et~al.}, ``Graph-based path planning
  for autonomous robotic exploration in subterranean environments,'' in
  \emph{International Conference on Intelligent Robots and Systems
  (IROS)}.\hskip 1em plus 0.5em minus 0.4em\relax IEEE, 2019, pp. 3105--3112.

\bibitem{batinovic2021multi}
A.~Batinovic, T.~Petrovic, A.~Ivanovic \emph{et~al.}, ``A multi-resolution
  frontier-based planner for autonomous 3{D} exploration,'' \emph{IEEE Robotics
  and Automation Letters}, vol.~6, no.~3, pp. 4528--4535, 2021.

\bibitem{kim2022autonomous}
H.~Kim, H.~Kim, S.~Lee \emph{et~al.}, ``Autonomous exploration in a cluttered
  environment for a mobile robot with {2D}-map segmentation and object
  detection,'' \emph{IEEE Robotics and Automation Letters}, vol.~7, no.~3, pp.
  6343--6350, 2022.

\bibitem{surmann2003autonomous}
H.~Surmann, A.~N{\"u}chter, and J.~Hertzberg, ``An autonomous mobile robot with
  a {3D} laser range finder for {3D} exploration and digitalization of indoor
  environments,'' \emph{Robotics and Autonomous Systems}, vol.~45, no.~3, pp.
  181--198, 2003.

\bibitem{bircher2016receding}
A.~Bircher, M.~Kamel, K.~Alexis \emph{et~al.}, ``Receding horizon
  ``next-best-view'' planner for {3D} exploration,'' in \emph{International
  Conference on Robotics and Automation (ICRA)}.\hskip 1em plus 0.5em minus
  0.4em\relax IEEE, 2016, pp. 1462--1468.

\bibitem{dai2020fast}
A.~Dai, S.~Papatheodorou, N.~Funk \emph{et~al.}, ``Fast frontier-based
  information-driven autonomous exploration with an {MAV},'' in
  \emph{International Conference on Robotics and Automation (ICRA)}.\hskip 1em
  plus 0.5em minus 0.4em\relax IEEE, 2020, pp. 9570--9576.

\bibitem{betz2022autonomous}
J.~Betz, H.~Zheng, A.~Liniger \emph{et~al.}, ``Autonomous vehicles on the edge:
  A survey on autonomous vehicle racing,'' \emph{IEEE Open Journal of
  Intelligent Transportation Systems}, vol.~3, pp. 458--488, 2022.

\bibitem{shrestha2019learned}
R.~Shrestha, F.-P. Tian, W.~Feng \emph{et~al.}, ``Learned map prediction for
  enhanced mobile robot exploration,'' in \emph{International Conference on
  Robotics and Automation (ICRA)}.\hskip 1em plus 0.5em minus 0.4em\relax IEEE,
  2019, pp. 1197--1204.

\bibitem{qiao2019sampling}
W.~Qiao, Z.~Fang, and B.~Si, ``A sampling-based multi-tree fusion algorithm for
  frontier detection,'' \emph{International Journal of Advanced Robotic
  Systems}, vol.~16, no.~4, p.~14, 2019.

\bibitem{seewald2022energy}
A.~Seewald, H.~Garc{\'i}a~de Marina, H.~S. Midtiby \emph{et~al.},
  ``Energy-aware planning-scheduling for autonomous aerial robots,'' in
  \emph{International Conference on Intelligent Robots and Systems
  (IROS)}.\hskip 1em plus 0.5em minus 0.4em\relax IEEE, 2022, pp. 2946--2953.

\bibitem{garcia2017guidance}
H.~Garc{\'i}a~de Marina, Y.~A. Kapitanyuk, M.~Bronz \emph{et~al.}, ``Guidance
  algorithm for smooth trajectory tracking of a fixed wing {UAV} flying in wind
  flows,'' in \emph{International Conference on Robotics and Automation
  (ICRA)}.\hskip 1em plus 0.5em minus 0.4em\relax IEEE, 2017, pp. 5740--5745.

\bibitem{seewaldphdthesis}
A.~Seewald, ``Energy-aware coverage planning and scheduling for autonomous
  aerial robots,'' Ph.D. thesis, Syddansk Universitet, 2021,
  {\tt\footnotesize\href{https://doi.org/10.21996/7ka6-r457}{doi.org/10.21996/7ka6-r457}}.

\bibitem{labbe2019rtab}
M.~Labb{\'e} and F.~Michaud, ``{RTAB-M}ap as an open-source {LiDAR} and visual
  simultaneous localization and mapping library for large-scale and long-term
  online operation,'' \emph{Journal of Field Robotics}, vol.~36, no.~2, pp.
  416--446, 2019.

\bibitem{setterfield2013terrain}
T.~P. Setterfield and A.~Ellery, ``Terrain response estimation using an
  instrumented rocker-bogie mobility system,'' \emph{IEEE Transactions on
  Robotics}, vol.~29, no.~1, pp. 172--188, 2013.

\bibitem{kirchgeorg2022multimodal}
S.~Kirchgeorg and S.~Mintchev, ``Multimodal aerial-tethered robot for tree
  canopy exploration,'' in \emph{International Conference on Intelligent Robots
  and Systems (IROS)}.\hskip 1em plus 0.5em minus 0.4em\relax IEEE, 2022, pp.
  6080--6086.

\bibitem{amster2020turtlebot}
R.~Amsters and P.~Slaets, ``{TurtleBot} 3 as a robotics education platform,''
  in \emph{International Conference on Robotics in Education (RiE)}.\hskip 1em
  plus 0.5em minus 0.4em\relax Springer, 2020, pp. 170--181.

\bibitem{goncalves2010vector}
V.~M. Goncalves, L.~C.~A. Pimenta, C.~A. Maia \emph{et~al.}, ``Vector fields
  for robot navigation along time-varying curves in $n$-dimensions,''
  \emph{IEEE Transactions on Robotics}, vol.~26, no.~4, pp. 647--659, 2010.

\bibitem{campos2021orb}
C.~Campos, R.~Elvira, J.~J.~G. Rodr{\'i}guez \emph{et~al.}, ``{ORB-SLAM3}: {A}n
  accurate open-source library for visual, visual-inertial, and multimap
  {SLAM},'' \emph{IEEE Transactions on Robotics}, vol.~37, no.~6, pp.
  1874--1890, 2021.

\bibitem{bickler1989articulated}
D.~B. Bickler, ``Articulated suspension system,'' {U.S. P}atent~4 840 394, June
  20, 1989.

\bibitem{schmid2020efficient}
L.~Schmid, M.~Pantic, R.~Khanna \emph{et~al.}, ``An efficient sampling-based
  method for online informative path planning in unknown environments,''
  \emph{IEEE Robotics and Automation Letters}, vol.~5, no.~2, pp. 1500--1507,
  2020.

\bibitem{seewald2019coarse}
A.~Seewald, U.~P. Schultz, E.~Ebeid \emph{et~al.}, ``Coarse-grained
  computation-oriented energy modeling for heterogeneous parallel embedded
  systems,'' \emph{International Journal of Parallel Programming}, vol.~49,
  no.~2, pp. 136--157, 2021.

\end{thebibliography}
}

\end{document}